\documentclass{article}

\PassOptionsToPackage{numbers, compress}{natbib}
\usepackage[preprint]{neurips_2026}   

\usepackage[utf8]{inputenc}
\usepackage[T1]{fontenc}
\usepackage{hyperref}
\usepackage{url}
\usepackage{microtype}
\usepackage{graphicx}
\usepackage{subcaption}
\usepackage{booktabs}

\usepackage{array}
\usepackage{multirow}
\newcolumntype{L}[1]{>{\raggedright\arraybackslash}m{#1}}
\newcolumntype{C}[1]{>{\centering\arraybackslash}m{#1}}

\usepackage{fontawesome5}

\usepackage{amsmath}
\usepackage{amssymb}
\usepackage{amsfonts}
\usepackage{mathtools}
\usepackage{amsthm}
\usepackage{nicefrac}
\usepackage{tabularx}
\usepackage{makecell}
\renewcommand{\arraystretch}{1.1}

\usepackage{placeins}
\usepackage{soul}
\usepackage{tikz}
\usepackage{pgf-pie}
\newcommand{\bcircrender}[1]{%
  \tikz[baseline=(char.base)]{%
    \node[shape=circle, fill=black, inner sep=1.2pt] (char)
      {\color{white}\scriptsize\bfseries #1};%
  }%
}
\DeclareRobustCommand{\bcirc}[1]{\texorpdfstring{\bcircrender{#1}}{(#1)}}
\usepackage{url}
\usepackage{algorithm}
\usepackage{algorithmic}

\usepackage[capitalize,noabbrev]{cleveref}
\usepackage[most]{tcolorbox}
\tcbuselibrary{breakable}
\usepackage{fvextra}
\usepackage{xcolor}
\usepackage{fontawesome5}
\theoremstyle{plain}

\theoremstyle{definition}

\theoremstyle{remark}

\usepackage[textsize=tiny]{todonotes}

\newcommand{\methodname}{ViDR}
\newcommand{\benchmarkname}{MMR Bench+}

\newtcolorbox{casebox}[1]{%
  enhanced, colback=blue!3, colframe=blue!40!black,
  fonttitle=\bfseries, title={#1},
  rounded corners, boxrule=0.8pt, breakable,
  top=6pt, bottom=6pt, left=6pt, right=6pt
}
\newtcolorbox{chainbox}[1]{%
  enhanced, colback=gray!6, colframe=gray!50,
  fonttitle=\bfseries\small, title={#1},
  rounded corners, boxrule=0.5pt,
  top=4pt, bottom=4pt, left=5pt, right=5pt,
  before skip=6pt, after skip=6pt
}

\newtcolorbox{taskrule}{%
  enhanced, breakable,
  colback=white,
  colframe=white,
  boxrule=0pt,
  borderline west={2pt}{0pt}{blue!40!black},
  left=8pt, right=0pt, top=3pt, bottom=3pt,
  before skip=2pt, after skip=8pt,
  fontupper=\small,
}

\title{ViDR: Grounding Multimodal Deep Research Reports in Source Visual Evidence}

\author{%
  Zhuofan Shi$^{1,2}$\thanks{Equal contribution.} \And
  Peilun Jia$^{3}$\textsuperscript{*} \And
  Baoqin Sun$^{1,2}$\textsuperscript{*} \AND
  Haiyang Shen$^{1,2}$ \And
  Sixiong Xie$^{1,2}$ \And
  Yun Ma$^{1,2}$ \AND
  Xiang Jing$^{1,2}$ \thanks{Corresponding author.}\\[0.5em]
  $^{1}$School of Software and Microelectronics, Peking University, Beijing 100871, China \\
  $^{2}$National Key Laboratory of Data Space Technology and System, Beijing 100871, China \\
  $^{3}$School of Software Engineering, Beijing Jiaotong University, Beijing, 100044, China \\
  $^{4}$College of Computer Science and Electronic Engineering, Hunan University, Changsha, 410082, China \\
  \faGithub~Code: \url{https://github.com/PKU-JX-LAB/PKU_MMDR}
}

\begin{document}

\maketitle

\begin{abstract}
Recent advances in deep research systems have significantly improved the ability of large language models (LLMs) to generate long, grounded reports through iterative retrieval and reasoning. On one hand, text centered deep research frameworks excel at multistep search and synthesis but largely operate over textual evidence. On the other hand, emerging multimodal deep research systems extend retrieval to images and charts, but they still predominantly either produce figure-free reports or synthesize charts themselves, rather than retrieving and integrating figures from the original sources. As a result, they fail to fully leverage source visual evidence. In this paper, we present \methodname{}, a multimodal deep research framework that grounds long-form reports in source figures. Rather than relying only on generated charts, \methodname{} treats source figures as evidence objects that can be retrieved, interpreted, routed, and verified, while still generating analytical charts when needed. \methodname{} builds an evidence-indexed outline that links report claims to textual and visual evidence, refines noisy web images into source-figure evidence atoms through context-aware filtering, outline-aware reranking, and VLM-based visual analysis, and generates each section with section-specific evidence. It further validates visual references to prevent hallucinated or misplaced figures, while generating analytical charts when necessary. We further introduce \benchmarkname{}, a benchmark for evaluating visual evidence use in deep research reports. Beyond conventional report quality and citation accuracy, \benchmarkname{} measures whether systems can retrieve, place, and interpret source visual artifacts, as well as generate analytical charts when needed. Experiments show that \methodname{} achieves stronger overall report quality and substantially improves source-figure integration and verifiability over strong commercial and open-source baselines. These results suggest that visual evidence should not be overlooked in multimodal deep research reports, as source figures can strengthen evidential grounding, improve visual support, and enhance report verifiability. 
\end{abstract}

\section{Introduction}

\begin{figure}[t]
  \centering
  \includegraphics[width=\linewidth]{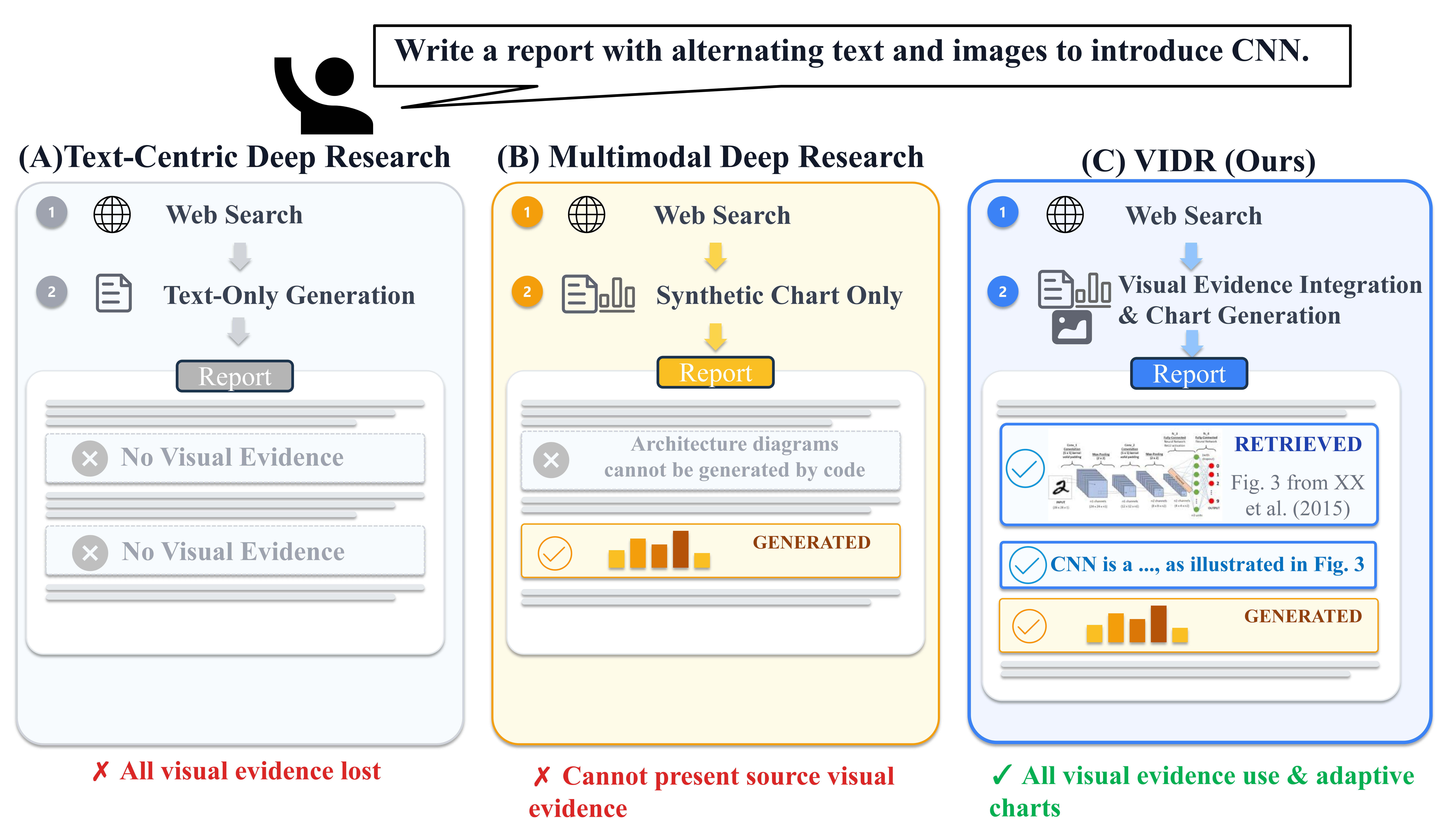}
    \caption{
    Comparison of deep research report paradigms. Text-centric systems omit visual evidence, while chart-generation systems synthesize plots but lose source figures. \methodname{} retrieves and grounds source figures in the report, generating analytical charts only when source evidence is insufficient.
    }
  \label{fig:motivation}
\end{figure}

Deep research agents~\cite{zhang2025deep,huang2025deep} have made long form research synthesis increasingly practical. Given an open ended question, these systems can plan multistep searches, browse the web, collect evidence, and produce structured reports. Commercial systems such as OpenAI Deep Research~\cite{openai_introducing_deep_research_2025}, Gemini Deep Research~\cite{google_gemini_deep_research_2024}, and Tongyi DeepResearch~\cite{team2025tongyi} show the practical value of this paradigm, while academic systems such as STORM~\cite{shao-etal-2024-assisting}, WebThinker~\cite{li2025webthinker}, WebDancer~\cite{wu2025webdancer}, and DualGraph~\cite{shi2026dualgraph} improve different parts of the research loop.

However, the final artifacts produced by most deep research agents remain largely text centered. This is a poor match for many research tasks. In scientific papers, technical reports, policy briefings, and market analysis, figures are often not decorative supplements. They are evidence: a model architecture diagram can define the mechanism being discussed, a training curve can substantiate a convergence claim, a benchmark table can support a comparative conclusion, and a map or schematic can explain a spatial or causal relationship. When a deep research agent discards these visual artifacts, it loses part of the source evidence and forces the reader to trust a textual paraphrase of information that was originally visual.

Existing approaches do not fully solve this problem. As shown in \Cref{fig:motivation}, text centered systems usually produce plain reports without embedded figures, even when relevant images or charts appear in the retrieved pages. Recent multimodal report systems, such as Multimodal DeepResearcher~\cite{yang2025multimodaldeepresearcher}, can generate charts from extracted data, which is useful for summarizing numerical evidence. Yet generated charts are not substitutes for original source figures. Source figures preserve provenance, visual details, labels, layouts, experimental context, and design choices that may be essential for interpreting the underlying claim. Replacing them with regenerated plots can remove exactly the evidence a reader needs to inspect.

Retrieving source figures for long reports is also not a simple matter of saving every image from a webpage. Web pages contain large numbers of logos, thumbnails, icons, social buttons, advertisements, and duplicated assets. Even after filtering obvious noise, a candidate figure must be judged against the research topic, connected to the right part of an evolving report outline, interpreted through its visible content, and inserted only where it can support the surrounding argument. The writing stage adds another challenge: a model may mention an image without analyzing it, cite an image ID that was never provided, or reuse the same figure across unrelated sections. A visually grounded report therefore requires the whole research and generation pipeline to preserve evidence structure, not only a better image crawler.

We propose \methodname{}, a deep research framework that treats source visual evidence as a first class component of report generation. During research, \methodname{} collects textual and visual evidence while maintaining an evolving outline whose claims cite supporting learning IDs. This outline steers later searches toward weakly supported sections and later becomes the evidence indexed report plan. Candidate images pass through context aware filtering, topic reranking with outline context, and VLM based visual analysis, producing compact evidence atoms that describe what is visible and what claim the figure can support. The final report is then generated section by section: each section receives only its routed textual evidence and local image candidates, and generated text must ground visual references in concrete landmarks such as axes, labels, trends, modules, colors, or spatial layout. When no suitable source figure exists, \methodname{} generates analytical charts from the collected evidence under iterative refinement.

We also introduce \benchmarkname{}, a visually grounded evaluation benchmark for deep research reports. It contains 160 bilingual research queries across 16 domains and evaluates both general report quality and the use of source visual evidence. In addition to informativeness, coherence, and verifiability, the benchmark measures visualization quality and source-figure integration, allowing us to distinguish systems that merely write fluent reports from systems that can retrieve, place, and interpret source figures.

Our experiments show that \methodname{} improves both textual and visual report quality. Compared with strong commercial and open source baselines, \methodname{} achieves the best overall score on \benchmarkname{} and the largest gains on source-figure integration. These results suggest that source-figure integration is not just a presentation feature. It improves the evidential basis of long form research reports and exposes a capability gap that text only evaluation fails to measure.

\section{Related Work}
\label{sec:related_work}
\subsection{Deep Research Agents}

Deep research agents automate the loop of planning, web search, evidence aggregation, and long form synthesis.
Early systems such as STORM~\cite{shao-etal-2024-assisting} demonstrated that LLMs can write Wikipedia style articles by simulating conversations from multiple perspectives and grounding claims in retrieved sources.
GPT-Researcher~\cite{feleovic_gpt_researcher_github} and LangChain Open Deep Research~\cite{langchain_open_deep_research_github} further operationalized this idea through open source agentic pipelines that decompose research queries into subquestions and conduct parallel web searches.
Commercial systems have made this paradigm practical: OpenAI Deep Research~\cite{openai_introducing_deep_research_2025} uses extended reasoning and iterative browsing to produce comprehensive reports, Gemini Deep Research~\cite{google_gemini_deep_research_2024} integrates Google's search infrastructure for multistep investigation, and Tongyi DeepResearch~\cite{team2025tongyi} introduces outline guided search and hierarchical report generation with citation grounding.
Academic systems improve different parts of the research loop: WebThinker~\cite{li2025webthinker} interleaves chain of thought reasoning with search actions; Search-o1~\cite{li2025search} and Search-R1~\cite{jin2025search} train LLMs to reason over search results through reinforcement learning; WebDancer~\cite{wu2025webdancer} formulates information seeking as an autonomous agency problem; WebWeaver~\cite{li2025webweaver} structures web scale evidence through dynamic outlines; DualGraph~\cite{shi2026dualgraph} separates knowledge exploration from outline structure via jointly evolving knowledge and outline graphs; and WebShaper~\cite{tao2025webshaper} studies agentic synthesis of training data through formalized information seeking.
Recent surveys~\cite{zhang2025deep,huang2025deep} provide systematic views of this rapidly growing landscape.

Despite this progress, most deep research agents remain text-centered at the report level. Recent multimodal agents such as WebWatcher~\cite{geng2025webwatcher} and
Vision-DeepResearch~\cite{huang2026visiondeepresearch} extend search to multimodal evidence, while Multimodal
DeepResearcher~\cite{yang2025multimodaldeepresearcher} interleaves reports with generated charts. In contrast, \methodname{} treats retrieved source figures as traceable evidence objects and routes them into section-level
long-form report generation.

\subsection{Multimodal Document and Report Generation}

A separate line of work studies how long documents can combine prose with visual artifacts. Long form text generation methods such as RECURRENTGPT~\cite{xiong2025beyond} improve planning and coherence, but typically treat the output as text only. Recent multimodal report systems introduce charts and visual elements into generated reports.
Multimodal DeepResearcher~\cite{yang2025multimodaldeepresearcher} generates reports that interleave text and charts through a four stage pipeline and uses Formal Description of Visualization specifications to generate diverse charts from extracted data.
Deep-Reporter~\cite{ye2026deepreporter} extends multimodal long report generation with multimodal search, checklist guided synthesis, and recurrent context management, while EvidFuse~\cite{lin2026evidfuse} studies writing time evidence construction for consistent reporting with text and charts in structured data settings.

These systems show that visual content can improve report synthesis, but their visual pipelines mainly rely on generated charts or auxiliary image context. \methodname{} instead treats original source figures as evidence objects, retrieving, enriching, and aligning them with report sections while generating analytical charts when source evidence is insufficient.

\subsection{Evaluation of Deep Research}

Evaluating autonomous research agents requires measuring multiple dimensions,
including evidence gathering, synthesis, factuality, citation accuracy, and
presentation. DeepResearch Bench~\cite{du2025deepresearch} introduces 100
PhD-level tasks and evaluates reports with the RACE framework and the FACT
citation metric. DeepResearch Bench II~\cite{li2026deepresearch} extends this
with expert-derived rubrics. DeepResearchGym~\cite{coelho2025deepresearchgym}
provides a reproducible evaluation sandbox, while
DeepScholar-Bench~\cite{patel2025deepscholar} supports live benchmarking of
generative research synthesis. MiroEval~\cite{ye2026miroeval} further broadens
evaluation to multimodal deep research through report-level quality assessment,
factuality verification, and process-centered audits. BrowseComp~\cite{wei2025browsecomp}
and GAIA~\cite{mialon2023gaia} evaluate complementary browsing and general
assistant capabilities.

Recent benchmarks also examine visual evidence in complex research tasks.
VDR-Bench~\cite{zeng2026visiondeepresearchbenchmarkrethinkingvisual} focuses on visual-textual search and multimodal
evidence retrieval. More closely related to multimodal deep research,
Multimodal DeepResearcher~\cite{yang2025multimodaldeepresearcher} introduces
MultimodalReportBench (MMR-Bench) for evaluating text-chart interleaved research
reports, while MMDeepResearch-Bench~\cite{huang2026mmdeepresearchbenchbenchmarkmultimodaldeep}
evaluates text--visual completeness and cross-modal grounding in multimodal
deep research agents. However, these benchmarks mainly assess visual evidence
retrieval, understanding, or grounding, with less emphasis on section-level placement and long-form report use. Our \benchmarkname{}
therefore evaluates whether source figures are relevant, coherent with the surrounding text, and used to support the report narrative.

\section{Method}
\label{sec:method}

\begin{figure}[t]
  \centering
  \includegraphics[width=\linewidth]{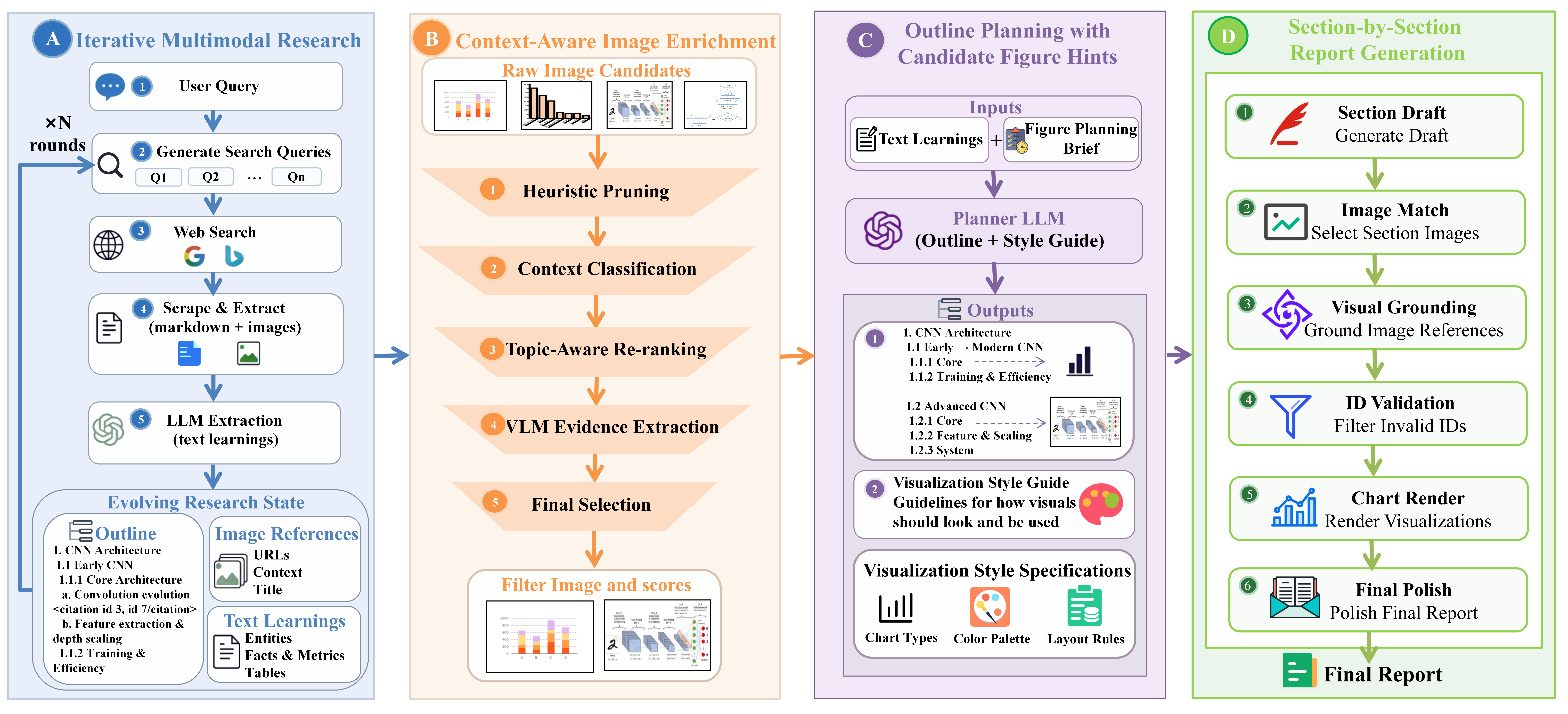}
  \caption{Overview of the \methodname{} pipeline: \bcirc{A}~multimodal research, \bcirc{B}~image enrichment, \bcirc{C}~evidence-indexed planning, and \bcirc{D}~grounded sectionwise generation.}
  \label{fig:framework}
\end{figure}

We present the \methodname{} framework, a four stage pipeline that enables deep research agents to retrieve, filter, organize, and integrate source visual evidence into generated reports. As illustrated in \Cref{fig:framework}, the pipeline proceeds through: \bcirc{A}~Outline Evolving Multimodal Research, \bcirc{B}~Context Aware Image Enrichment, \bcirc{C}~Evidence Indexed Report Planning, and \bcirc{D}~Sectionwise Report Generation with Visual Grounding Verification.

\subsection{Stage \texorpdfstring{\bcirc{A}}{A}: Outline Evolving Multimodal Research}
\label{sec:stage_a}

This stageconstructs a multimodal evidence pool $\mathcal{E}$ and an evolving outline $\mathcal{O}_t$ over $N_R$ search rounds to steer later research. Given a topic $q$, the agent retrieves textual and visual web evidence, extracts compact factual learnings, and updates $\mathcal{O}_t$ by linking supported claims to learning IDs and marking under-evidenced regions as \emph{Gap} entries. Conditioned on $\mathcal{O}_t$, subsequent queries target these gaps, making the outline both a search controller and an evidence index for downstream generation.

\paragraph{Iterative Evidence Acquisition.}
At round $t$, the agent generates $N_K$ search queries, retrieves the top $N_P$ pages for each query, and extracts page markdown together with embedded image references. The resulting evidence is added to a growing multimodal pool $\mathcal{E}=\mathcal{E}_{\text{text}}\cup\mathcal{E}_{\text{visual}}$.
Textual evidence is used for learning extraction, while visual references are preserved as candidate source figures for downstream enrichment and placement.

\paragraph{Extraction-Time Image Filtering.}
Web pages often contain non-evidential images such as logos, icons, thumbnails, advertisements, and social-media assets. Passing all such images to later VLM stages would waste context and increase the risk of irrelevant figure insertion. \methodname{} therefore applies a conservative extraction-time filter $f_{\mathrm{pre}}$ that removes visually non-evidential web assets using metadata-level cues, producing a tractable candidate set for subsequent semantic and visual reasoning. This filter does not decide final report usage; final selection is deferred to the context-aware enrichment stage. Implementation details, including thresholds and blocklists, are provided in
\Cref{app:implementation}.

\paragraph{Learning Extraction with Image Preservation.}
Each retrieved page is distilled by an LLM into structured \emph{learnings}:
compact factual summaries that retain key entities, dates, metrics, tables, and source links. Each learning receives a persistent identifier $L_i$ so that later stages can cite and route evidence at the claim level. Unlike text-only summarization, our extraction step preserves visual references. Images encountered during scraping are represented as structured placeholders, e.g., \texttt{[Image\_X: description]}, which the LLM is instructed to retain in the extracted learnings. For each referenced image, the model also produces a \emph{deductive evidence atom} describing the visible features, the supported factual claim, and the rationale connecting the image to the research topic. This keeps source figures linked to their supported claims, allowing later stages to treat images as evidence rather than unstructured media.

\paragraph{Online Outline Maintenance.}
Once nonempty evidence is obtained, the agent initializes a hierarchical outline whose fine-grained claims cite supporting learning IDs through explicit citation tags, e.g., \texttt{<citation>id 3, id 7</citation>}. After each subsequent search round, new evidence is attached to relevant claims, the structure is revised when needed, and under-supported parts are marked as \emph{Gap} entries. The resulting outline jointly represents the intended report structure and the current evidence coverage.

The updated outline then guides the next round of query generation, together with the research direction, the latest learnings, and previously issued queries. By exposing both supported claims and evidence gaps, the outline reduces redundant search and directs later rounds toward under-evidenced report regions.

\subsection{Stage \texorpdfstring{\bcirc{B}}{B}: Context Aware Image Enrichment Pipeline}
\label{sec:stage_b}

The raw visual pool from Stage~\bcirc{A} is inherently noisy: web pages often mix informative figures with decorative assets, and even valid images may be only tangentially related to the research topic. Forwarding all candidates to the report generator would waste context budget and increase the risk of irrelevant or misattributed visual references. \methodname{} therefore applies a funnel-shaped enrichment pipeline that distills noisy web images into a compact set of source figures with explicit visual evidence annotations.

\paragraph{Heuristic Pruning.}
The pipeline first applies an asymmetric rule-based filter to remove obvious web assets while preserving ambiguous candidates. Candidates are pruned when their metadata or local context contains asset signals such as \texttt{logo} or \texttt{avatar}, unless they also exhibit technical signals such as \texttt{figure} or \texttt{table}. This conservative design preserves recall for downstream semantic and visual filtering.

\paragraph{Context Classification.}
Each remaining candidate is classified using its surrounding webpage context. The classifier predicts whether the image should be kept, assigns a figure type, estimates contextual relevance, and flags likely primary-source images. Decorative, noisy, or irrelevant candidates are removed before topic-level reranking.

\paragraph{Topic Reranking with Outline Context.}
The surviving candidates are reranked by jointly considering the research topic, accumulated learnings, the current outline $\mathcal{O}_t$, and the candidate's local textual context. The ranker assigns each image a topical relevance score and a recommended report section aligned with the current outline. Conditioning on $\mathcal{O}_t$ keeps image selection consistent with the evolving report structure and avoids placing images under stale or hallucinated section headings.

\paragraph{VLM-Based Visual Analysis.}
Because surrounding text may not faithfully describe the actual figure, \methodname{} next applies a VLM to inspect top-ranked candidates directly. The VLM extracts visible titles, key textual elements, and a concise visual summary. It also produces \emph{deductive evidence atoms} that connect visible features to supported factual claims and rationales, linking each visual feature to a supported claim and the rationale for that connection. These atoms provide concrete visual anchors for evidence-grounded report generation.

\paragraph{Final Selection and Score Merging.}
Finally, an LLM aggregates the accumulated signals for each visually analyzed candidate and produces a use/reject decision, a composite score, supported claims, and integration advice. Selected images are summarized into compact \emph{planning briefs} with figure types, evidence atoms, and recommended
placements for downstream planning and generation.

\subsection{Stage \texorpdfstring{\bcirc{C}}{C}: Evidence Indexed Report Planning}
\label{sec:stage_c}

After research and image enrichment, \methodname{} converts the evolving research outline into an executable report plan. Rather than regenerating the report structure from scratch, Stage~\bcirc{C} preserves the citation and
\emph{Gap} metadata accumulated during Stage~\bcirc{A}. The internal outline is kept evidence-indexed for routing, while a cleaned version is exposed as the reader-facing outline. This separation allows the report structure to remain grounded in the research process while still supporting a polished external
presentation.

\paragraph{Research Outline Finalization.}
The outline $\mathcal{O}=\{s_1,s_2,\ldots,s_K\}$ is represented as a numbered hierarchy, where top-level sections define generation units and lower-level bullets encode concrete claims. Fine-grained bullets contain \texttt{<citation>} tags that point to learning IDs, while \emph{Gap} entries mark claims or subtopics with insufficient support. Before generation, \methodname{} finalizes $\mathcal{O}$ as the authoritative section structure and removes citation and \emph{Gap} metadata only from the reader-facing outline.

\paragraph{Section Evidence Indexing.}
For internal generation, \methodname{} parses $\mathcal{O}$ and aggregates the citation IDs under each top-level section. This yields a section-level evidence map $R(s_k)=\{L_i\}$ and an optional gap annotation $G(s_k)$. The map constrains section writing to the evidence cited within the corresponding outline branch: section $s_k$ receives only the learning items indexed by $R(s_k)$. Evidence-sparse sections are explicitly marked rather than filled with unrelated global context.

\paragraph{Style and Visual Planning.}
Stage~\bcirc{C} also produces a visualization style guide for downstream section generation. Image candidates have already been aligned to the report outline through Stage~\bcirc{B}'s topic reranking. Final image usage is resolved dynamically during Stage~\bcirc{D}; here, the planner specifies report tone, figure style, and chart design constraints to ensure visual consistency across sections.

\subsection{Stage \texorpdfstring{\bcirc{D}}{D}: Sectionwise Report Generation}
\label{sec:stage_d}

Stage~\bcirc{D} turns the evidence-indexed plan into the final report. Rather than generating the report in a single LLM call, \methodname{} iterates over the top-level sections in $\mathcal{O}$. For each section, it routes local textual and visual evidence, generates grounded content under explicit media-use rules, and validates the resulting visual references. After all sections are assembled, report-level media placeholders are resolved and the report is polished while
preserving verified visual elements.

\paragraph{Section Evidence Routing.} For each section $s_k$, textual evidence is selected using the citation map $R(s_k)$ extracted from the research outline. The generator receives only the learning items cited within the corresponding outline branch, rather than the entire evidence pool. If the section has a gap annotation $G(s_k)$, the prompt exposes this limitation and instructs the model to write conservatively instead of filling the gap with unsupported claims. This local routing makes each section auditable and reduces interference from evidence intended for other sections.

\paragraph{Section Image Routing.} Visual evidence is routed locally in the same manner. Given the section title, summary, and routed learnings, \methodname{} reranks the enriched image pool using image summaries, evidence roles, nearby text, keywords, source credibility, and figure type. Images such as architecture diagrams, pipelines, schematics, result charts, benchmarks, and ablation tables are prioritized when they match the section context. A deduplication key tracks selected source figures to avoid reusing the same visual evidence across sections.

\paragraph{Grounded Section Generation.} The section prompt contains the routed learnings, selected image candidates, the visualization style guide from Stage~\bcirc{C}, and explicit media-grounding rules. When the model inserts a source-figure placeholder such as \texttt{[Image\_X]}, it must state the claim supported by the image and refer to concrete visual landmarks, such as labels, trends, module names, colors, or spatial layout. The prompt permits a \texttt{<visualization>} specification only when the section contains real data or a concrete mechanism that benefits from a generated chart or schematic. If prose or retrieved source figures already convey the point, the model is instructed not to generate additional media.

\paragraph{Section Reference Validation.} After each section is generated, \methodname{} validates its visual references before proceeding to the next section. Image placeholders that were not provided to the section are removed, preventing hallucinated references from entering the final report. Repeated references to the same image are collapsed, and the used image set is updated so that later sections avoid duplicates. This validation step enforces local visual grounding at the section level.

\paragraph{Report-Level Media Resolution and Polishing.} Once all sections are assembled, \methodname{} resolves media placeholders at the report level. Valid \texttt{<visualization>} specifications are converted into D3 HTML charts through an actor-critic refinement loop: the actor implements the chart, the critic checks for issues such as placeholder data, unreadable labels, excessive margins, and style violations, and revised candidates are selected by pairwise comparison~\cite{yang2025multimodaldeepresearcher}. Source image placeholders are then replaced with their source-figure URLs. Finally, a Write Agent polishes the complete report for organization, coherence, and verifiability. To preserve media during rewriting, all images and rendered charts are temporarily replaced with opaque anchors and restored afterward, ensuring that visual elements are neither deleted nor duplicated.

\subsection{\benchmarkname{}: Visually-Grounded Evaluation Benchmark}
\label{sec:benchmark}

Existing benchmarks for deep research systems focus on textual quality, citation accuracy, and factual correctness, but overlook a critical requirement of multimodal deep research: the ability to retrieve, select, and integrate source visual evidence in a way that materially supports the narrative. To address this limitation, we construct \benchmarkname{} for evaluating visually grounded deep research reports. Building upon MMR-Bench~\cite{yang2025multimodaldeepresearcher} proposed by Yang et al., we extend the benchmark in two ways: (1)~broadening the query set to cover a wider range of domains and topic complexities, and (2)~redesigning the evaluation dimensions to explicitly measure whether visual evidence is used effectively rather than merely inserted.

\paragraph{Task Construction.}
\benchmarkname{} comprises \textbf{160 bilingual research queries} spanning 16 domains across science, technology, policy, and public affairs (\Cref{fig:benchmark_domains}). Among them, 100 English queries are adopted from MMR-Bench~\cite{yang2025multimodaldeepresearcher}, and 60 Chinese queries are newly curated with domain expertise following the procedure in
\Cref{app:query_construction}. Each query asks the system to produce a complete deep research report with textual analysis and supporting visual evidence. The newly curated queries broaden the original English set and introduce more challenging \emph{Visually Demanding Queries} (VDQ), where source figures from papers or web pages are useful for explaining methods, results, mechanisms, or comparisons. This design allows \benchmarkname{} to evaluate both general
report-generation ability and scenarios where visual evidence is central to the topic, without reducing the benchmark to an image-insertion task. The bilingual design broadens topic coverage across languages. The evaluation dimensions are summarized in \S\ref{para:eval_dimensions}, with full scoring rubrics provided
in \Cref{app:eval_rubric}.

\section{Experiments}
\label{sec:experiments}

\subsection{Experimental Setup}
We evaluate all systems on \benchmarkname{}, the benchmark introduced in \Cref{sec:benchmark}.

\paragraph{Evaluation Dimensions.}
\label{para:eval_dimensions}
Each report is scored from 1 to 5 along five dimensions: Informativeness and Depth, Coherence and Organization, Verifiability, Visualization Quality, and Source-Figure Integration. These dimensions jointly evaluate textual quality, evidential grounding, visual clarity, and whether source figures are selected, placed, and interpreted appropriately. Detailed rubrics are provided in \Cref{app:eval_rubric}.

\paragraph{Baselines.}
We compare \methodname{} against both closed source commercial deep research systems (Gemini-DR, Baidu Qianfan DR-agent, Qwen-Deep-Research, OpenAI-DR) and open source frameworks (Tongyi-DR, WebShaper, Multimodal DeepResearcher~\cite{yang2025multimodaldeepresearcher}). 

\paragraph{Implementation Details.}
We instantiate \methodname{} with GPT-5.2 and Gemini-3-Flash as backbone models, and use GPT-5.2 as the judge. The default budget is capped at 5 search rounds, 10 queries per round, and 3 retrieved pages per query. \methodname{} generates reports section by section under evolving outline guidance. 

\subsection{Main Results}
\label{sec:main_results}
\begin{table}[!t]
    \caption{\label{tab:deepresearch_bench}
      Main results on \benchmarkname{}. Higher is better.
      Abbreviations: DR = DeepResearch; MMDR = Multimodal DeepResearcher; Info. = Informativeness; Coher. = Coherence; Verif. = Verifiability; Viz.Q. = Visualization Quality; Src.Fig. = Source-Figure Integration; Ovrl. = Overall.
      Open source baselines use the backbone model shown in parentheses when applicable.
    }
  \centering
  \footnotesize
  \setlength{\tabcolsep}{3pt}
  \renewcommand{\arraystretch}{0.95}
  \resizebox{\columnwidth}{!}{%
  \begin{tabular}{>{\raggedright\arraybackslash}m{3.5cm}cccccc}
    \toprule
    \textbf{Agent Systems} &
    \textbf{Info.} &
    \textbf{Coher.} &
    \textbf{Verif.} &
    \textbf{Viz.Q.} &
    \textbf{Src.Fig.} &
    \textbf{Ovrl.} \\
    \midrule
    \multicolumn{7}{c}{\textit{Closed Source}} \\
    Gemini-DR~\citep{google_gemini_deep_research}         & $3.97\pm0.02$ & $3.71\pm0.01$ & $3.09\pm0.03$ & $1.72\pm0.03$ & $1.02\pm0.03$ & $2.70\pm0.01$ \\
    Baidu Qianfan DR-agent      & $3.82\pm0.01$ & $3.87\pm0.05$ & $2.73\pm0.02$ & $1.34\pm0.03$ & $1.00\pm0.00$ & $2.55\pm0.02$ \\
    Qwen-Deep-Research          & $3.98\pm0.01$ & $3.93\pm0.02$ & $2.84\pm0.03$ & $1.82\pm0.02$ & $1.00\pm0.00$ & $2.71\pm0.00$ \\
    OpenAI-DR~\citep{openai_introducing_deep_research_2025}                   & $4.03\pm0.02$ & $3.86\pm0.03$ & $3.48\pm0.05$ & $1.71\pm0.03$ & $1.00\pm0.00$ & $2.82\pm0.01$ \\
    \midrule
    \multicolumn{7}{c}{\textit{Open Source}} \\
    Tongyi-DR(30B)~\citep{team2025tongyi}         & $3.31\pm0.02$ & $3.64\pm0.01$ & $2.29\pm0.01$ & $1.60\pm0.02$ & $1.04\pm0.01$ & $2.38\pm0.01$ \\
    WebShaper~\citep{tao2025webshaper}         & $3.13\pm0.02$ & $3.50\pm0.02$ & $2.75\pm0.01$ & $1.23\pm0.02$ & $1.05\pm0.01$ & $2.33\pm0.01$ \\
    
    MMDR(GPT-5.2)~\cite{yang2025multimodaldeepresearcher}
    & $3.44\pm0.02$ & $3.89\pm0.02$ & $3.03\pm0.01$ & $3.62\pm0.01$ & $1.00\pm0.00$ & $3.00\pm0.00$ \\

    \midrule
    \textbf{\methodname{} (GPT-5.2)}     & $4.15\pm0.01$ & $4.00\pm0.03$ & $4.03\pm0.01$ & $3.85\pm0.03$ & $3.66\pm0.04$ & $3.94\pm0.02$ \\
    \textbf{\methodname{} (Gemini-3-Flash)} & $4.09\pm0.04$ & $3.66\pm0.04$ & $3.53\pm0.03$ & $3.97\pm0.03$ & $3.42\pm0.08$ & $3.74\pm0.03$ \\
    \bottomrule
  \end{tabular}}
\end{table}
\Cref{tab:deepresearch_bench} presents the main results. \methodname{} achieves the best overall score of $3.94$, outperforming the strongest closed source system, OpenAI-DR ($2.82$), and the strongest prior multimodal baseline, Multimodal DeepResearcher ($3.00$). The improvement is not confined to visual metrics. \methodname{} also obtains the highest informativeness ($4.15$), coherence ($4.00$), and verifiability ($4.03$), suggesting that source-figure retrieval and section-level evidence routing improve report quality rather than merely adding visual material.

The most direct gain appears in source-figure integration. Existing deep
research systems score close to the floor value of $1.00$ on Src.Fig. because
they rarely provide traceable source figures, making section relevance, claim
support, and visual grounding difficult to assess. Multimodal DeepResearcher
achieves strong visualization quality ($3.62$) through generated charts, but its
Src.Fig. score remains $1.00$. In contrast, \methodname{} reaches $3.66$ on
Src.Fig. and the highest visualization quality score ($3.85$), showing that retrieved source figures improve evidence support beyond chart quality alone.

These results also clarify why text only evaluation is insufficient for multimodal deep research. Tongyi-DR and several closed source systems produce coherent reports, but their visual evidence scores remain low. \benchmarkname{} therefore exposes a capability gap that is hidden by standard report quality metrics: whether an agent can find, place, and interpret source visual evidence inside a long form report.

\subsection{Ablation Study}
\label{sec:ablation}

To isolate the contribution of the two key components in \methodname{}, we conduct two targeted ablation experiments on \benchmarkname{}. All variants reuse the same Stage~\bcirc{A} search outputs to ensure a fair comparison. Results are reported in \Cref{tab:ablation}.

\begin{table}[h]
  \centering
  \caption{\label{tab:ablation}
    Ablation study on \benchmarkname{}. Each row removes one key component from the full \methodname{} pipeline. 
  }
  \footnotesize
  \setlength{\tabcolsep}{3pt}
  \renewcommand{\arraystretch}{0.95}
  \resizebox{\columnwidth}{!}{%
  \begin{tabular}{>{\raggedright\arraybackslash}m{3.2cm}cccccc}
    \toprule
    \textbf{Variant} & \textbf{Info.} & \textbf{Coher.} & \textbf{Verif.} & \textbf{Viz.Q.} & \textbf{Src.Fig.} & \textbf{Ovrl.} \\
    \midrule
    \methodname{} (full)                      & $4.15\pm0.01$ & $4.00\pm0.03$ & $4.03\pm0.01$ & $3.85\pm0.03$ & $3.66\pm0.04$ & $3.94\pm0.02$ \\
    w/o Context Aware Image Enrichment         & $4.14\pm0.04$ & $3.88\pm0.02$ & $3.89\pm0.03$ & $3.86\pm0.03$ & $3.16\pm0.08$ & $3.78\pm0.02$ \\
    w/o Sectionwise Generation          & $3.98\pm0.03$ & $3.77\pm0.03$ & $3.68\pm0.04$ & $3.90\pm0.02$ & $3.46\pm0.07$ & $3.76\pm0.02$ \\
    \bottomrule
  \end{tabular}}
\end{table}

\paragraph{w/o Context Aware Image Enrichment.}
This variant preserves the Stage~\bcirc{A} image crawling and the final image insertion pipeline, but disables the context aware image enrichment process in Stage~\bcirc{B}. Instead of using the full multimodal filtering, reranking, and visual understanding pipeline, the system relies only on weak textual metadata associated with each image for image selection. This ablation tests whether context aware image enrichment is necessary for identifying and integrating relevant source figures.

\paragraph{w/o Sectionwise Generation.}
This variant replaces the sectionwise drafting process in Stage~\bcirc{D} with a single global report generation pass. The model writes the full report from the pooled research evidence and selected image candidates at once, without evidence routing by section or visual grounding during each section draft. This ablation tests whether sectionwise decomposition improves writing quality in long contexts, coherence, verifiability, and image integration.

\paragraph{Analysis.}
As shown in \Cref{tab:ablation}, both components improve \methodname{} in complementary ways. Removing context aware image enrichment causes the largest drop in source-figure integration, from $3.66$ to $3.16$. This suggests that the image enrichment module not only improves figure selection, but also strengthens alignment between images and text and evidence grounding. Without rich visual evidence signals, the system is less able to interpret and use image content, weakening the evidential support of the report. Removing sectionwise generation leads to broader declines in informativeness, coherence, and verifiability, reducing the overall score from $3.94$ to $3.76$. This indicates that sectionwise generation improves content organization, evidence coverage, and factual consistency by routing evidence to the appropriate sections.

\section{Limitations}
\label{sec:discussion}
\methodname{} depends on the availability and quality of source figures on the open web, and image selection can remain difficult when webpage context is sparse or ambiguous. The pipeline also introduces additional cost due to iterative search, image enrichment, sectionwise generation, and report-level refinement. Future work may improve figure-to-claim alignment, complex document parsing.

\section{Conclusion}
\label{sec:conclusion}
We presented \methodname{}, a deep research framework that retrieves, grounds, and integrates source visual evidence into long-form report generation. 
Together with \benchmarkname{}, which evaluates both evidence grounding and analytical visualization, our experiments show that \methodname{} improves textual and visual report quality, especially in source-figure integration. 
The results suggest that source figures and grounded analytical visualizations can strengthen the evidential basis of deep research reports, while showing that text-only evaluation may overlook important aspects of multimodal evidence use in report generation.

\newpage

\bibliography{custom}
\bibliographystyle{plainnat}

\newpage
\appendix

\section{Query Construction Process}
\label{app:query_construction}
\begin{figure}[t]
    \centering
    \begin{tikzpicture}
    \pie[
        radius=2.8, text=legend, sum=auto,
        color={cyan!60, orange!60, green!50, yellow!60, red!40, blue!40, purple!40, pink!50, teal!50, brown!40, cyan!35, orange!35, green!35, red!30, blue!30, purple!30},
        before number=, after number=,
    ]{
        15/Tech \& Media,
        15/Healthcare,
        15/AI \& ML,
        14/Climate \& Env,
        13/Agri \& Food,
        13/Economy \& Work,
        11/Econ \& Policy,
        10/Systems \& HW,
        10/Biomed \& Health,
        10/Energy \& Climate,
        9/Energy,
        8/Population,
        6/Education,
        4/Travel,
        4/Industry \& Platforms,
        3/Public Sector
    }
    \end{tikzpicture}
    \caption{Domain distribution of the 160 research queries in \benchmarkname{}, spanning 16 domains across science, technology, policy, and public affairs.}
    \label{fig:benchmark_domains}
\end{figure}
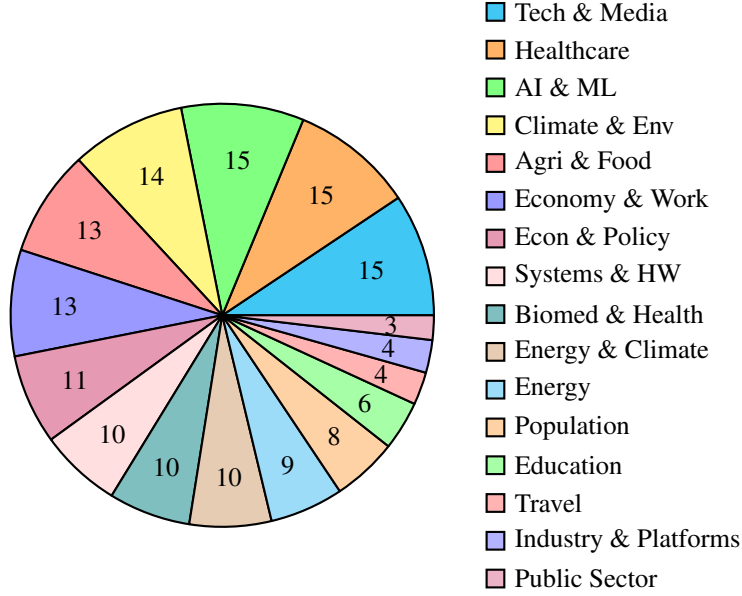

The 160 research queries in \benchmarkname{} are constructed through a systematic curation process designed to ensure domain diversity, open-ended research difficulty, and varying levels of visual-evidence demand.

\paragraph{Query Sources and Curation.}
The 100 English queries are adopted from
MMR-Bench~\cite{yang2025multimodaldeepresearcher} to preserve comparability
with prior work. These queries cover open-ended research topics that require
substantive synthesis across sources rather than single-document lookup. The
additional 60 Chinese queries are curated with domain expertise to broaden the
benchmark beyond the original English set and to cover more challenging topics
with diverse visual-evidence requirements.

For the newly curated Chinese queries, we follow a three-step construction
process. First, we select candidate topics from recent research papers, technical reports, public policy discussions, and high-impact news events, with emphasis on topics that require multi-source evidence synthesis. Second, we rewrite these topics into open-ended research queries that ask for a complete deep research report rather than a short factual answer. Third, we review the candidate queries for clarity, scope, domain coverage, and answerability, revising or removing queries that are ambiguous, overly narrow, duplicative, or unlikely to have sufficient public evidence.

\paragraph{Visual-Evidence Demand Characterization.}
The newly curated Chinese queries include a subset of more challenging topics
with higher visual-evidence demand, referred to as \emph{Visually Demanding
Queries} (VDQ). These topics often involve model architectures, training
dynamics, ablation studies, system comparisons, or scientific mechanisms, where figures from papers or web pages are useful for faithfully explaining methods, results, and comparisons. For each VDQ, we verify that relevant public sources with informative visual evidence, such as architecture diagrams, benchmark tables, training curves, or official diagrams, are accessible online. This step ensures that visually demanding queries are grounded in retrievable evidence rather than hypothetical targets. VDQ is used for subset analysis and difficulty characterization, while all queries are evaluated using the same report-level rubric described in \Cref{app:eval_rubric}.

\paragraph{Quality Control.}
All queries undergo a final quality review to ensure that: (i)~no duplicate or
near-duplicate topics exist across the English and Chinese subsets; (ii)~the 16 target domains are approximately balanced (see \Cref{fig:benchmark_domains}); and (iii)~each query is self-contained and does not presuppose domain-specific jargon without sufficient context.

\section{Implementation Details}
\label{app:implementation}

\paragraph{Backbone Models and API Configuration.}
We instantiate \methodname{} under two backbone configurations. In the GPT configuration, the research, planning, global report generation, and section writing stages use GPT-5.2, whereas visual analysis, chart generation, and chart critique use GPT-5.4-mini. In the Gemini configuration, all generation side stages use Gemini-3-Flash. Unless otherwise stated, generation calls are made with a temperature of 0.7 and a timeout of 300 seconds. We use GPT-5.2 as the judge model for report evaluation.

\paragraph{Research Stage Configuration.}
The outline evolving multimodal research stage (\S\ref{sec:stage_a}) runs for $N_R = 5$ search iterations, with $N_K = 10$ queries per iteration and $N_P = 3$ pages retrieved per query. Each
query produces up to $N_L = 3$ learnings and two follow-up questions. Scraped
page content is truncated to 8{,}000 characters before learning extraction. The
extraction-time pre-filter $f_{\mathrm{pre}}$ requires a minimum file size of
30KB, excludes \texttt{.svg}, \texttt{.gif}, and \texttt{data:image} formats,
validates HTTP content type when available, and applies filename, alt-text, and
domain-level blocklists. The per-page extraction cap is set to
$N_{\mathrm{cap}} = 15$ images, and the surrounding text window size is
$W = 140$ characters.

\paragraph{Image Enrichment Pipeline.}
The five stage enrichment pipeline (\S\ref{sec:stage_b}) uses the following capacity limits: up to 100 images survive context classification (batch size 12), 50 pass topic reranking informed by the outline, 20 undergo VLM based visual analysis, and 15 are retained after final selection.

\paragraph{Dynamic Evidence Routing.}
The image ranker for each section (\S\ref{sec:stage_d}) scores each candidate image for a given section $s_k$ by:
\begin{equation}
\label{eq:image_score}
\text{score}(v_i, s_k) = \underbrace{\alpha \cdot |\text{tok}(s_k) \cap \text{tok}(v_i)|}_{\text{keyword overlap}} + \underbrace{c_i}_{\text{credibility}} + \underbrace{\beta_{\text{type}}}_{\text{figure bonus}} + \underbrace{\gamma \cdot \mathbf{1}[|r_i| > \tau]}_{\text{depth bonus}}
\end{equation}
where $\text{tok}(\cdot)$ denotes tokenization, $c_i$ is the credibility score of image $v_i$ derived from metadata, and $r_i$ denotes the length of its evidence role description. In our implementation, $\alpha=2.0$, $\beta_{\text{type}} \in \{+3.0$ for architecture, pipeline, hardware, and schematic figures; $+2.0$ for result charts, benchmark figures, and ablation tables$\}$, $\gamma=1.0$, and $\tau=20$ characters. Images with fewer than 2 keyword matches are discarded before scoring, and only candidates with $\text{score}(v_i, s_k) > 4.0$ are retained.

\paragraph{Chart Generation.}
Charts are generated as self contained D3.js (v7) HTML files rendered via a headless browser (Selenium with Microsoft Edge). The actor critic loop runs for up to 3 iterations. A template detection heuristic rejects charts containing placeholder signals (e.g., ``monthly performance,'' ``sample data,'' or $\geq$4 month names indicating a calendar template) and forces regeneration.

\paragraph{Report-Level Refinement.}
The final report-level refinement pass truncates the report to 90{,}000 characters and the learnings to 45{,}000 characters before sending them to the LLM. Media elements are replaced with opaque tokens (\texttt{[[MEDIA\_ANCHOR\_001]]}); after refinement, a validation step checks bijective preservation of all tokens. If any anchor is missing or duplicated, a targeted repair prompt restores it with minimal textual changes. The refinement pass also cleans residual editorial scaffolding, normalizes heading numbering, and appends an automatically extracted references section from markdown links and arXiv identifiers found in both the report and the learnings.

\paragraph{Prompt Templates and Runtime.}
We release the key prompt templates in the supplementary codebase, including those for query generation and learning extraction, outline planning, section writing, chart actor critic refinement, and final report rewriting. All experiments use hosted API services for LLM inference and a standard workstation for orchestration and chart rendering.

\section{Evaluation Rubrics}
\label{app:eval_rubric}

We evaluate each report along five dimensions that jointly capture content
quality, structural coherence, evidential grounding, visualization quality, and
source-figure integration. Each dimension is scored from 1 to 5 with 0.5-point
increments using a structured LLM-as-judge rubric. Unless otherwise stated, the
overall score is computed as the average of the five dimension scores.

A key design choice in our rubric is to distinguish general visualization
quality from source-figure integration. \emph{Visualization Quality} evaluates
the quality of visible visual content, including clarity, readability, labeling,
and communicative effectiveness, regardless of whether the visual content is
retrieved or generated. In contrast, \emph{Source-Figure Integration} evaluates
whether retrieved source figures are appropriately selected, credible, and
effectively integrated with the report narrative. This dimension is not an
image-count metric: high scores require source figures to support important
claims and be clearly interpreted in the surrounding text. In particular, reports
without explicitly provided traceable source figures receive a score of 1 on the
Source-Figure Integration dimension, since the judge cannot assess source-figure
relevance, credibility, or integration without source visual evidence.

\begin{itemize}
  \item \textbf{Informativeness and Depth (Info.)}: Whether the report provides
  comprehensive, substantive, and sufficiently detailed analysis of the target
  topic. High-scoring reports go beyond surface-level description and offer
  meaningful explanation, synthesis, and contextualization.

  \item \textbf{Coherence and Organization (Coher.)}: Whether the report follows
  a clear and coherent narrative structure with logically organized sections,
  smooth transitions, and visual elements integrated in a way that supports
  rather than disrupts the narrative flow.

  \item \textbf{Verifiability (Verif.)}: Whether major claims, numerical
  statements, and analytical conclusions are grounded in credible and traceable
  evidence. Reports receive higher scores when readers can clearly identify the
  evidential basis of the main arguments.

  \item \textbf{Visualization Quality (Viz.Q.)}: The quality of the visible
  visual content itself, including clarity, readability, labeling, and
  communicative effectiveness. High-scoring visualizations improve reader
  understanding rather than serve as decorative elements. This dimension evaluates
  real visible visuals, including retrieved source figures and generated charts,
  but does not give credit for placeholders, code, Markdown tables, captions, or
  textual descriptions alone.

  \item \textbf{Source-Figure Integration (Src.Fig.)}: Whether retrieved source
  figures are appropriately selected, credible, and effectively integrated with
  the report narrative. High-scoring reports use source figures to support
  important claims and provide clear interpretation in the surrounding text. This
  dimension assesses whether visual evidence is integral to the report narrative
  rather than peripheral.
\end{itemize}

\section{Evaluation Reliability Checks}
\label{app:eval_reliability}

Because our evaluation relies on LLM-as-judge scoring, we perform two reliability
checks. First, we measure the retest stability of the same judge under an
identical evaluation setup. Second, we conduct a human audit to assess whether
the automatic scores and rationales are aligned with human inspection. Together,
these checks help characterize whether the reported trends are stable and
interpretable under both repeated automatic scoring and human review.

\subsection{Stability of Repeated LLM-as-Judge Scoring}
\label{app:judge_reliability}

We randomly sample a subset of reports generated by multiple baseline systems and
evaluate each report three times independently using the same judge and rubric.
For each evaluation dimension, we report three stability metrics:
Within-0.5 Agreement, Mean Range, and Mean Std. Within-0.5 Agreement denotes the
percentage of reports for which the maximum score difference across the three
runs is no larger than 0.5 points. Mean Range and Mean Std measure the average
dispersion of repeated scores.

\begin{table}[t]
  \centering
  \caption{
    Stability of repeated LLM-as-judge scoring on a randomly sampled subset of
    reports from multiple baselines. Higher is better for Within-0.5 Agreement,
    while lower is better for Mean Range and Mean Std.
  }
  \label{tab:judge_stability_subset}
  \small
  \setlength{\tabcolsep}{6pt}
  \begin{tabular}{lccc}
    \toprule
    \textbf{Dimension} & \textbf{Within-0.5 (\%)} & \textbf{Mean Range} & \textbf{Mean Std} \\
    \midrule
    Informativeness and Depth     & 100.00 & 0.1167 & 0.0674 \\
    Coherence and Organization    & 99.17  & 0.1625 & 0.0938 \\
    Verifiability                 & 96.67  & 0.2625 & 0.1490 \\
    Visualization Quality         & 95.83  & 0.2292 & 0.1310 \\
    Source-Figure Integration     & 95.83  & 0.1167 & 0.0655 \\
    Overall                       & 99.17  & 0.1375 & 0.0741 \\
    \bottomrule
  \end{tabular}
\end{table}

As shown in \Cref{tab:judge_stability_subset}, repeated LLM-as-judge scoring is
stable across all evaluation dimensions. Within-0.5 Agreement exceeds 95\% for
every dimension, and both Mean Range and Mean Std remain small. Verifiability and
Visualization Quality exhibit slightly higher variance than the other
dimensions, which is expected because they depend more on factual attribution and
visual interpretation. Overall, these results indicate that the observed
evaluation trends are not primarily driven by random instability in repeated
judge scoring.

\subsection{Human Evaluation of LLM-as-Judge Scores}
\label{app:judge_sanity_check}

To assess whether the automatic scores align with human inspection, we conduct a
human evaluation of LLM-as-judge scores on a representative subset of
\benchmarkname{}. The subset is selected to cover both the original MMR-Bench
queries and Visually Demanding Queries (VDQ), as well as multiple domains in the
benchmark. For each sample, three annotators are shown the query, the generated
report, the provided source figures, if any, and the automatic judge score with
its rationale for each evaluation dimension. Annotators then judge whether each
automatic score is \emph{reasonable}, \emph{partially reasonable}, or
\emph{unreasonable}. System identities are hidden from annotators.

This evaluation examines whether the structured LLM-as-judge rubric produces
interpretable and defensible scores under human review. For the Source-Figure
Integration dimension, annotators specifically verify whether the score reflects
the relevance, credibility, and textual integration of source figures, rather
than merely the number of inserted images. We define the acceptable rate as the
sum of the reasonable and partially reasonable rates. The percentages in
\Cref{tab:judge_sanity_check} are computed over all annotator--sample judgments
for each dimension.

\paragraph{Interpretation of partially reasonable judgments.}
A judgment is marked as \emph{partially reasonable} when the automatic score is
mostly defensible but the accompanying rationale is incomplete or insufficiently
specific. This category indicates partial agreement with the judge decision
rather than disagreement with the overall scoring direction.

\begin{table}[t]
\centering
\caption{
Human evaluation of LLM-as-judge scores. Three annotators judge whether each automatic score and rationale is reasonable, partially reasonable, or unreasonable for each evaluation dimension. Accept. = Reason. + Part. Reason.
}
\label{tab:judge_sanity_check}
\small
\setlength{\tabcolsep}{5pt}
\begin{tabular}{lcccc}
\toprule
\textbf{Dimension} & \textbf{Reason. (\%)} & \textbf{Part. Reason. (\%)} & \textbf{Unreason. (\%)} & \textbf{Accept. (\%)} \\
\midrule
Informativeness & 77.1 & 19.8 & 3.1 & 96.9 \\
Coherence & 74.0 & 21.9 & 4.2 & 95.8 \\
Verifiability & 69.8 & 22.9 & 7.3 & 92.7 \\
Visualization Quality & 64.6 & 26.0 & 9.4 & 90.6 \\
Source-Figure Integration & 60.4 & 29.2 & 10.4 & 89.6 \\
\midrule
Average & 69.2 & 24.0 & 6.9 & 93.1 \\
\bottomrule
\end{tabular}
\end{table}

As shown in \Cref{tab:judge_sanity_check}, the average acceptable rate across the five evaluation dimensions reaches 93.1\%, and every dimension exceeds 89\%. Source-Figure Integration has the lowest reasonable rate, reflecting the greater difficulty of judging visual-evidence integration, but it still achieves an acceptable rate of 89.6\%. These results indicate that the automatic scores
are broadly consistent with human review, including on the most visually grounded evaluation dimension.

\section{Performance on General and Visually Demanding Queries}
\label{app:vdq_subset}

We further analyze \methodname{} on two query groups: the original MMR-Bench
queries, which reflect general long-form research tasks, and the newly curated
Visually Demanding Queries (VDQ), which place stronger emphasis on visual
evidence. The VDQ subset is more challenging because these topics require not
only textual synthesis, but also retrieval and interpretation of source visual
evidence to explain methods, results, mechanisms, or comparisons. Therefore, its
scores are expected to be lower than those on the general MMR-Bench subset.

As shown in \Cref{tab:vdq_subset}, \methodname{} achieves an overall score of
$4.04$ on the MMR-Bench subset and $3.77$ on the VDQ subset. The lower VDQ
score reflects the additional difficulty of visually grounded evidence
integration, especially in verifiability and visualization-related dimensions.
Nevertheless, the VDQ subset still receives a Src.Fig. score of $3.58$,
suggesting that \methodname{} can ground visual evidence in the report narrative
even under more demanding visual-evidence conditions.

\begin{table}[t]
\centering
\caption{
Performance on general and visually demanding queries. MMR-Bench denotes the
original general-query subset, while VDQ denotes the newly curated visually
demanding query subset. Scores are reported as mean $\pm$ standard error. Higher
is better.
}
\label{tab:vdq_subset}
\small
\setlength{\tabcolsep}{4pt}
\begin{tabular}{lcccccc}
\toprule
\textbf{Subset} & \textbf{Info.} & \textbf{Coher.} & \textbf{Verif.} & \textbf{Viz.Q.} & \textbf{Src.Fig.} & \textbf{Overall} \\
\midrule
MMR-Bench
& $4.24{\pm}0.02$
& $4.05{\pm}0.01$
& $4.23{\pm}0.01$
& $3.95{\pm}0.05$
& $3.71{\pm}0.05$
& $4.04{\pm}0.01$ \\
VDQ
& $3.99{\pm}0.02$
& $3.92{\pm}0.08$
& $3.69{\pm}0.02$
& $3.69{\pm}0.05$
& $3.58{\pm}0.07$
& $3.77{\pm}0.03$ \\
\bottomrule
\end{tabular}
\end{table}

\section{Example Report Output}
\label{app:example_report}
\begin{figure}[h]
  \centering
  \includegraphics[width=\linewidth]{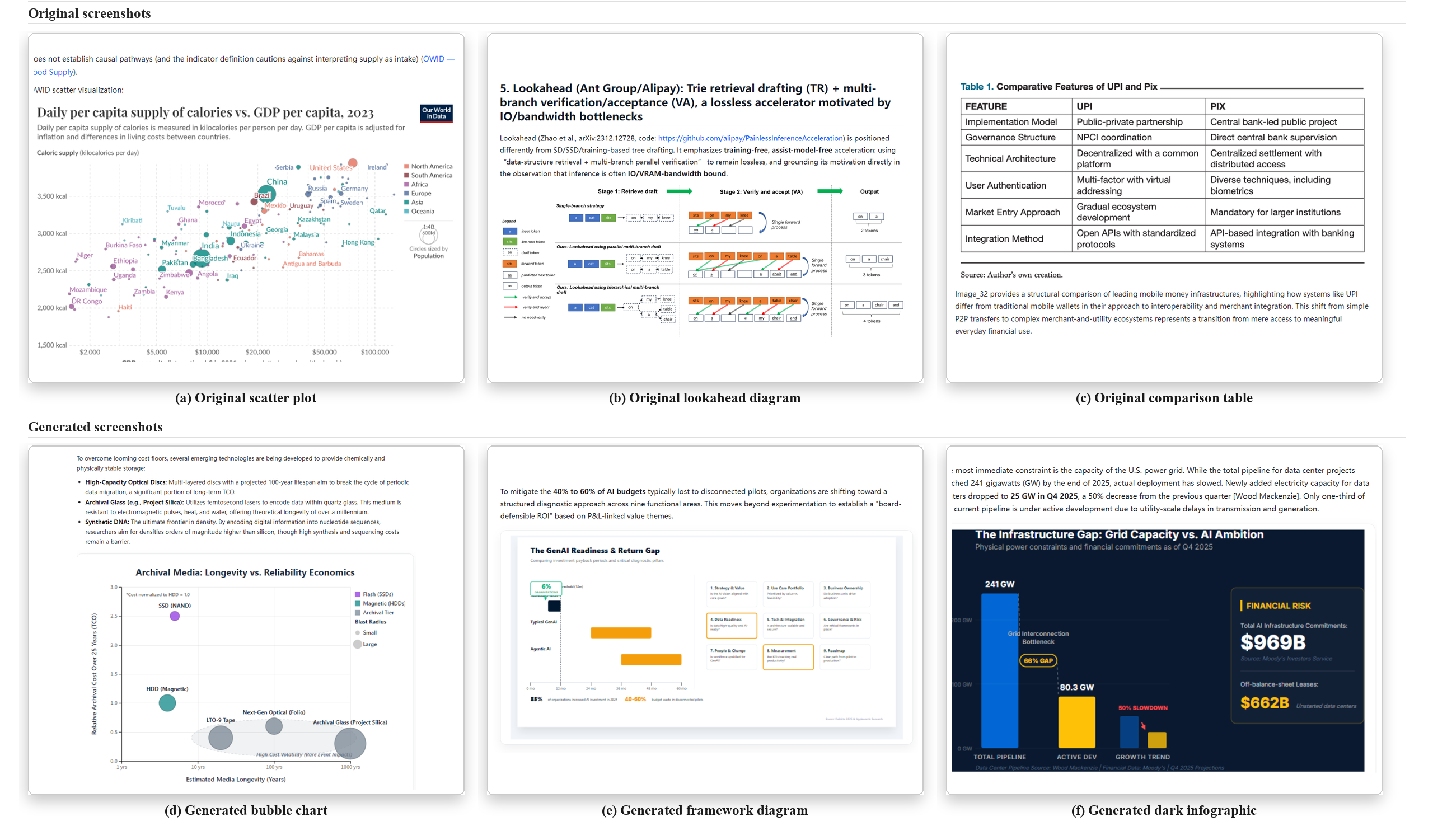}
  \caption{An excerpt from a report generated by \methodname{}. The report interleaves retrieved source figures with textual analysis, demonstrating fine grained alignment between visual evidence and report statements.}
  \label{fig:example_report}
\end{figure}

To illustrate the visually grounded report generation capability of \methodname{}, we present a representative excerpt from a system generated report in \Cref{fig:example_report}. Unlike conventional deep research systems that produce plain text outputs, \methodname{} generates reports with text and figures interleaved: retrieved source figures are embedded at contextually appropriate positions, each accompanied by descriptive captions that connect the visual evidence to the surrounding narrative. This example shows how retrieved source figures, rather than regenerated plots, are integrated to support the report's claims, yielding a more faithful and interpretable research output.

\section{Prompt Templates}
\label{app:prompts}

We provide representative prompt templates for the main LLM-mediated stages in
\methodname{}. For space, repeated formatting constraints and long schemas are
condensed, while the task instructions that define the behavior of each stage are
preserved. Deterministic parsing, evidence indexing, deduplication, and reference
validation steps are omitted because they are implemented procedurally rather than
through standalone prompt templates.

\subsection{Stage \texorpdfstring{\bcirc{A}}{A}: Initial Search Query Generation}
\label{app:prompt_initial_query}
\begin{tcolorbox}[
  colback=green!5!white,
  colframe=green!80!black,
  boxrule=0.8pt,
  rounded corners=all,
  arc=3pt,
  colbacktitle=green!90!black,
  coltitle=white,
  title={\quad \textsc{Prompt Template: Initial Search Query Generation}},
  fonttitle=\sffamily,
  fontupper=\rmfamily\scriptsize,
  colupper=black!85,
  breakable,
]
\begin{Verbatim}[breaklines,breakanywhere,commandchars=\\\{\}]
\textbf{System:}
You are an expert researcher. Be detailed, value good arguments over authority,
and consider recent or emerging developments when the user asks about them.

\textbf{User:}
Given the following prompt from the user, generate a list of SERP queries to
research the topic. Return a maximum of \{queries_num\} queries.
Make sure each query is unique and not similar to the others.

To find high-quality visual evidence, include visual-focused keywords such as
"architecture diagram", "performance benchmark chart", "comparison table", or
"visual schematic" in at least half of the generated queries.

OUTPUT FORMAT: Return only the queries, one per line.

<prompt>\{query\}</prompt>
Previous learnings:
\{learning_str\}
\end{Verbatim}
\end{tcolorbox}

\subsection{Stage \texorpdfstring{\bcirc{A}}{A}: Learning Extraction with Image Preservation}
\label{app:prompt_learning}
\begin{tcolorbox}[
  colback=green!5!white,
  colframe=green!80!black,
  boxrule=0.8pt,
  rounded corners=all,
  arc=3pt,
  colbacktitle=green!90!black,
  coltitle=white,
  title={\quad \textsc{Prompt Template: Learning Extraction}},
  fonttitle=\sffamily,
  fontupper=\rmfamily\scriptsize,
  colupper=black!85,
  breakable,
]
\begin{Verbatim}[breaklines,breakanywhere,commandchars=\\\{\}]
\textbf{System:}
You are an expert researcher extracting information from web pages.

\textbf{User:}
Given the following contents from a SERP search for <query>\{query\}</query>,
generate up to \{learning_num\} concise, detailed, information-dense learnings.
Include Markdown hyperlinks, named entities, exact metrics, dates, and useful
tables or lists.

If the contents contain image placeholders in the format [Image_X: description],
strictly preserve them in the extracted learnings. For each referenced image,
produce a deductive_evidence_atom with:
1. Visual Features: what is explicitly visible in the image.
2. Deductive Fact: the factual or quantitative claim supported by the image.
3. Rationale: how the image acts as evidence for the broader topic.

Also return up to \{question_num\} follow-up questions for further research.

<contents>\{contents\}</contents>
\end{Verbatim}
\end{tcolorbox}

\subsection{Stage \texorpdfstring{\bcirc{A}}{A}: Adaptive Outline Update and Query Guidance}
\label{app:prompt_adaptive_outline}
\begin{tcolorbox}[
  colback=green!5!white,
  colframe=green!80!black,
  boxrule=0.8pt,
  rounded corners=all,
  arc=3pt,
  colbacktitle=green!90!black,
  coltitle=white,
  title={\quad \textsc{Prompt Template: Adaptive Outline and Query Guidance}},
  fonttitle=\sffamily,
  fontupper=\rmfamily\scriptsize,
  colupper=black!85,
  breakable,
]
\begin{Verbatim}[breaklines,breakanywhere,commandchars=\\\{\}]
\textbf{Outline update system:}
You are an expert research planner. Maintain a living Markdown outline that
evolves as new evidence arrives. Update it by assigning new learning IDs to
claims via <citation> tags, restructuring sections when new subtopics appear,
and recording remaining evidence gaps.

\textbf{Outline update user:}
Topic: \{topic\}
Current outline:
\{current_outline\}
New learnings from round \{round_num\}:
\{new_learnings_str\}
All learnings so far:
\{all_learnings_str\}

Return only:
<adaptive_outline>
# 1. Section Title
## 1.1 Subsection Title
### 1.1.1 Sub-subsection Title
a. Specific point <citation>id 1, id 3</citation>
Gap: evidence still needed, if applicable
</adaptive_outline>

\textbf{Outline-guided query user:}
Given the current outline with evidence coverage and gaps, generate
\{queries_num\} targeted search queries for the next research round. Prioritize
Gap entries and weakly supported sections, avoid previously used queries, and
include visual-focused keywords in at least half of the queries.
\end{Verbatim}
\end{tcolorbox}

\subsection{Stage \texorpdfstring{\bcirc{B}}{B}: Topic Reranking with Outline Context}
\label{app:prompt_image_rerank}
\begin{tcolorbox}[
  colback=blue!5!white,
  colframe=blue!80!black,
  boxrule=0.8pt,
  rounded corners=all,
  arc=3pt,
  colbacktitle=blue!90!black,
  coltitle=white,
  title={\quad \textsc{Prompt Template: Outline-Aware Image Reranking}},
  fonttitle=\sffamily,
  fontupper=\rmfamily\scriptsize,
  colupper=black!85,
  breakable,
]
\begin{Verbatim}[breaklines,breakanywhere,commandchars=\\\{\}]
\textbf{System:}
You are an expert research assistant selecting source images for a report topic.
Use the topic and learnings to decide which candidate images are worth keeping.
Return strict JSON only.

\textbf{User:}
Topic:
\{topic\}

Report Outline:
\{outline_clean\}

Learnings:
\{trimmed_learnings\}

Rank these image candidates for relevance to the topic and align each useful
candidate with an actual section title from the outline.

Return a JSON array. Each item must be:
{
  "image_id": "...",
  "relevance_score": 0-5,
  "should_keep": true,
  "recommended_section": "...",
  "why_relevant": "..."
}

Candidates:
\{candidates_json\}
\end{Verbatim}
\end{tcolorbox}

\subsection{Stage \texorpdfstring{\bcirc{B}}{B}: VLM-Based Visual Analysis}
\label{app:prompt_vlm_analysis}
\begin{tcolorbox}[
  colback=blue!5!white,
  colframe=blue!80!black,
  boxrule=0.8pt,
  rounded corners=all,
  arc=3pt,
  colbacktitle=blue!90!black,
  coltitle=white,
  title={\quad \textsc{Prompt Template: VLM Visual Analysis}},
  fonttitle=\sffamily,
  fontupper=\rmfamily\scriptsize,
  colupper=black!85,
  breakable,
]
\begin{Verbatim}[breaklines,breakanywhere,commandchars=\\\{\}]
\textbf{System:}
You are an OCR and technical-figure understanding assistant.
Read visible text from the figure and extract structured visual evidence.
Return plain text with exactly these fields:
visible_title: ...
visible_text: ...
ocr_keywords: keyword1, keyword2, ...
ocr_summary: one sentence
deductive_evidence_atoms: Extract 3-5 key evidence atoms. For each atom, provide
[Visual Feature] -> [Deductive Fact] -> [Rationale].

\textbf{User:}
Topic: \{topic\}
Read this image carefully and extract OCR-like signals and evidence atoms.
\end{Verbatim}
\end{tcolorbox}

\subsection{Stage \texorpdfstring{\bcirc{C}}{C}: Visualization Style Guide Generation}
\label{app:prompt_style_guide}
\begin{tcolorbox}[
  colback=orange!5!white,
  colframe=orange!80!black,
  boxrule=0.8pt,
  rounded corners=all,
  arc=3pt,
  colbacktitle=orange!90!black,
  coltitle=white,
  title={\quad \textsc{Prompt Template: Visualization Style Guide}},
  fonttitle=\sffamily,
  fontupper=\rmfamily\scriptsize,
  colupper=black!85,
  breakable,
]
\begin{Verbatim}[breaklines,breakanywhere,commandchars=\\\{\}]
\textbf{System:}
You are an expert report-generation assistant specialized in creating
professional documents that combine insightful analysis with visualizations.

\textbf{User:}
Using the provided topic and previous learnings, produce a visualization style
guide for downstream section generation. In the default pipeline, the report
structure is inherited from the adaptive research outline; this prompt mainly
provides chart tone, color usage, information hierarchy, and figure style.

Topic: \{topic\}
Previous learnings: \{learning_str\}
Candidate figure evidence, if available: \{figure_plan_text\}

Return:
<style_guide>
[visualization style guide here]
</style_guide>
\end{Verbatim}
\end{tcolorbox}

\subsection{Stage \texorpdfstring{\bcirc{D}}{D}: Grounded Section Generation}
\label{app:prompt_section_generation}
\begin{tcolorbox}[
  colback=red!5!white,
  colframe=red!80!black,
  boxrule=0.8pt,
  rounded corners=all,
  arc=3pt,
  colbacktitle=red!90!black,
  coltitle=white,
  title={\quad \textsc{Prompt Template: Grounded Section Generation}},
  fonttitle=\sffamily,
  fontupper=\rmfamily\scriptsize,
  colupper=black!85,
  breakable,
]
\begin{Verbatim}[breaklines,breakanywhere,commandchars=\\\{\}]
You are writing a section for a research report on the topic: "\{topic\}".
This section is titled: "\{section_title\}".
Summary/Outline of this section: "\{section_summary\}".

Use the Visualization Style Guide:
\{style_guide\}

Use the following routed research learnings as your knowledge base:
\{sec_learnings\}

Known evidence gap for this section, if any:
\{evidence_gap\}

Available Image Placeholders for this section:
\{sec_images\}

Citation and verifiability rules:
- For every important quantitative claim, benchmark, chronology, causal
  explanation, comparison, or empirical judgment, attach a supporting Markdown
  hyperlink in the same sentence or immediately following sentence.
- Prefer source links already present in the routed learnings, especially primary
  or directly relevant sources.
- Do not rely on a single citation at the end of a long paragraph to support many
  distinct claims.
- If a claim is only partially supported, narrow or soften it instead of stating
  it strongly.
- If you compare systems, methods, periods, or regions, make sure each side of
  the comparison has nearby evidence.
- Keep the section auditable on its own.

Image grounding rules:
- Only insert an image if it directly supports a concrete claim in this section.
- Do not use weakly related or decorative images.
- Use only the provided image IDs; never make up fake image IDs.
- Do not use the same image multiple times in the report.
- Every inserted image must be explicitly explained in the surrounding text.
- If you insert an source figure, reference 1-2 concrete visual landmarks such
  as labels, trends, module names, colors, or spatial layout.
- Usually insert no more than 1-2 source figures in a section.

Generated visualization rules:
- Generate a <visualization> block only when real data or a grounded mechanism
  from the routed learnings supports it.
- Every visual element must come from the routed learnings or section context.
- Do not use dummy or fake data such as "Alpha", "Beta", or "Sample Bar Chart".
- If prose or available original source images already communicate the point
  clearly, do not generate an additional chart.
- A <visualization> block must contain a complete JSON design specification and
  must be closed immediately with </visualization>.

Write the full Markdown content for this section now.
\end{Verbatim}
\end{tcolorbox}

\subsection{Stage \texorpdfstring{\bcirc{D}}{D}: Chart Generation and Critique}
\label{app:prompt_chart_generation}
\begin{tcolorbox}[
  colback=red!5!white,
  colframe=red!80!black,
  boxrule=0.8pt,
  rounded corners=all,
  arc=3pt,
  colbacktitle=red!90!black,
  coltitle=white,
  title={\quad \textsc{Prompt Template: Chart Actor and Critic}},
  fonttitle=\sffamily,
  fontupper=\rmfamily\scriptsize,
  colupper=black!85,
  breakable,
]
\begin{Verbatim}[breaklines,breakanywhere,commandchars=\\\{\}]
\textbf{Chart actor system:}
You are a HTML/D3.js V7 implementation expert who transforms visualization
designs into working code.

\textbf{Chart actor user:}
Implement the visualization design specification with HTML and D3.js.
Treat the specification as binding. Do not replace it with a generic template.
Use titles, labels, data, captions, and semantic structure from the specification.
Import D3.js using <script src="https://d3js.org/d3.v7.min.js"></script>.
Set the root SVG/container width to exactly 700px, use sufficient margins, and
return a complete self-contained HTML file in a ```html code block.

\textbf{Chart critic user:}
Given a screenshot of the rendered HTML and console messages, check for:
design-spec mismatch, placeholder content, excessive annotations, overlapping
elements, sizing problems, excessive margins, and unreadable labels. For each
issue, describe the problem, its location, and the relevant elements. If no
issues are found, end with "No issues found."
\end{Verbatim}
\end{tcolorbox}
\subsection{Stage \texorpdfstring{\bcirc{D}}{D}: Report-Level Refinement with Media-Anchor Preservation}
\label{app:prompt_rewrite}
\begin{tcolorbox}[
  colback=red!5!white,
  colframe=red!80!black,
  boxrule=0.8pt,
  rounded corners=all,
  arc=3pt,
  colbacktitle=red!90!black,
  coltitle=white,
  title={\quad \textsc{Prompt Template: Report-Level Refinement}},
  fonttitle=\sffamily,
  fontupper=\rmfamily\scriptsize,
  colupper=black!85,
  breakable,
]
\begin{Verbatim}[breaklines,breakanywhere,commandchars=\\\{\}]
\textbf{System:}
You are an expert research report rewriter. Rewrite an existing Markdown report
into a clearer, better-organized, and better-supported research report.

Goals:
1. Improve structure, expression, analytical presentation, and readability.
2. Improve evidential support, precision, credibility, and verifiability.

Important rules:
- Use only information already present in the report, supplied learnings, and
  media inventory. Do not invent facts, data, sources, or claims.
- Preserve every [[MEDIA_ANCHOR_xxx]] token exactly once.
- Keep links, source names, citations, footnote-style references, and media anchors.
- Place source support near important quantitative claims, comparisons,
  mechanism explanations, and empirical judgments.
- If a claim is only partially supported, narrow it or add a caveat.
- Return only the final rewritten Markdown.

\textbf{User:}
Rewrite the following report into a stronger research report while preserving
all media anchors exactly once.

Topic: \{topic_hint\}
Media inventory: \{media_inventory\}
Learnings: \{learnings_text\}
Current report with media anchors: \{report_with_anchors\}
\end{Verbatim}
\end{tcolorbox}
\subsection{Evaluation: LLM-as-Judge Single-Report Scoring}
\label{app:prompt_eval_single}

\begin{tcolorbox}[
  colback=violet!5!white,
  colframe=violet!80!black,
  boxrule=0.8pt,
  rounded corners=all,
  arc=3pt,
  colbacktitle=violet!90!black,
  coltitle=white,
  title={\quad \textsc{Prompt Template: Single-Report Evaluation}},
  fonttitle=\sffamily,
  fontupper=\rmfamily\scriptsize,
  colupper=black!85,
  breakable,
]
\begin{Verbatim}[breaklines,breakanywhere,commandchars=\\\{\}]
\textbf{System:}
You are an expert evaluator of AI-generated reports with advanced knowledge of data visualization and information analysis.
Your role is to provide fair, impartial assessments of report quality based strictly on objective criteria.

## Evaluation Task
You will evaluate one AI-generated report based on:
- The overarching topic
- The report itself
- The provided visual inputs, if any

For each criterion below, assign a score from 1 to 5 (1=poor, 5=excellent) with half-point increments allowed (e.g., 3.5).
Provide a concise, evidence-based justification for each score. Be cautious with extreme scores (1 and 5).

## Additional Scoring Rules
- You must score visual dimensions primarily based on actually visible image content provided to you.
- Do not treat figure titles, captions, image placeholders, Markdown tables, Mermaid source code, or textual descriptions as equivalent to real rendered images.
- If a report does not contain real visible images, Visualization Quality should be scored conservatively and usually should not exceed 2.
- In these reports, <HTMLRenderer htmlFile="..."/> is a valid embedded-figure marker used by the report system. The corresponding rendered chart images may be provided separately in the evaluation input. Do not treat these tags by themselves as broken placeholders, missing content, or unfinished report artifacts.
- If the evaluation input already provides the corresponding real visible images, do not penalize Coherence and Organization merely because the markdown text contains <HTMLRenderer htmlFile="..."/> tags.
- When judging repetition, distinguish between harmful repetition and necessary analytical reinforcement. Briefly restating a key limitation, uncertainty, or methodological caveat in a later section should not be penalized unless it becomes near-duplicate, verbose, or materially disrupts flow.

## Mandatory Rule for Source-Figure Integration
- When scoring Source-Figure Integration, you must rely only on the images explicitly provided as source images in the evaluation input.
- Do not infer the existence of original/source images from report text alone.
- Do not treat figure titles, captions, links, citations, textual descriptions, Markdown image placeholders, Mermaid code, Mermaid renderings, generated charts, rendered screenshots, HTML screenshots, or model-generated diagrams as source/original images for this criterion.
- If no source/original images are explicitly provided in the evaluation input, Source-Figure Integration must be scored as 1.
- If the report claims to use source figures but no corresponding source/original images are actually provided, treat this as lack of visible evidence rather than successful integration.

## Evaluation Criteria
### Informativeness and Depth: Does the report deliver comprehensive, substantive and thorough information?
Score 1: Extremely superficial content with minimal information. Contains only basic facts without context or explanation.
Score 2: Limited content with some relevant information but significant gaps. Lacks necessary depth on key aspects.
Score 3: Adequate information covering main points with some supporting details, but missing opportunities for deeper analysis.
Score 4: Comprehensive information with substantive details, examples, and insights across most sections.
Score 5: Exceptionally thorough coverage with rich, nuanced details, expert-level insights, and well-contextualized information throughout.

### Coherence and Organization: Is the report well-organized with visualizations that connect meaningfully to the text?
Score 1: Disorganized; lacks logical structure and coherence. Visualizations appear random and unconnected to text.
Score 2: Basic structure is present but transitions are awkward, sections feel loosely assembled, or visuals are only weakly connected to the surrounding discussion.
Score 3: Clear overall organization with occasional flow issues, local repetition, or uneven transitions. Visualizations generally support the text, though integration or explanation could be tighter.
Score 4: Well-structured with smooth transitions between sections. Visualizations are meaningfully integrated with the text, and minor repetition or system-specific rendering markers do not materially disrupt reading.
Score 5: Impeccable organization with seamless progression of sections. Visualizations are tightly woven into the narrative, and the report maintains strong flow without unnecessary redundancy.

### Verifiability: Are the report's claims well supported with citations, references, or source annotations?
Score 1: Rarely supported with evidence; many claims are unsubstantiated.
Score 2: Inconsistently verified; some claims are supported; evidence is occasionally provided.
Score 3: Generally verified; claims are usually supported with evidence; however, there might be a few instances where verification is lacking.
Score 4: Well-supported; claims are very well supported with credible evidence, and instances of unsupported claims are rare.
Score 5: Very well-supported; almost every claim is substantiated with credible evidence, showing a high level of thorough verification.

### Visualization Quality: Is the visualization quality high? Score this mainly based on real visible image content, not chart titles, captions, placeholders, Markdown tables, Mermaid source code, or textual descriptions alone.
Score 1: No real visible images are provided, or visualization is severely missing; only placeholders, code, tables, or textual descriptions are present; or the charts are poor, confusing, misleading, badly labeled, or inappropriate.
Score 2: Visualization is extremely limited, or only a few simple or low-quality charts are provided; annotations and explanations are limited; axis, color, clarity, or layout problems clearly hinder understanding. If there are no real images and only tables or captions, the score should usually fall here.
Score 3: The charts are basically clear, have labels and annotations, and communicate information, but are not refined or miss opportunities for better expression. A score of 3 should only be given when real visible images exist and their quality is at least acceptable.
Score 4: The charts are well executed, visually effective, clearly labeled, reasonably annotated, thoughtfully designed, and the real visible images work well with the text.
Score 5: The charts are expert-level, highly polished, visually excellent, and reveal insights effectively. A score of 5 requires multiple high-quality, real visible, well-rendered images.

### Source-Figure Integration: Are the inserted original/source images (for example original paper figures, original charts, official diagrams, or source screenshots) selected appropriately, from credible sources, and effectively integrated with the text?
Score this criterion only based on the source/original images explicitly provided in the evaluation input.
Score 1: No source/original images are explicitly provided; or the provided source/original images are clearly irrelevant, unclear, low-quality, from questionable sources, or not meaningfully explained in the text.
Score 2: A small number of source/original images are provided, but their relevance is limited, their quality or source value is modest, or the text mentions them only briefly without forming meaningful support.
Score 3: The provided source/original images are basically relevant and reasonably credible, and the report offers some explanation, but the image selection or integration remains ordinary.
Score 4: The provided source/original images are well chosen, credible, and representative; the text explicitly cites and explains them, and they clearly support important points in the report.
Score 5: Multiple highly relevant, authoritative, and representative source/original images are provided; the report interprets them accurately and deeply, and they are tightly integrated into the argument in a way that substantially strengthens the report's persuasiveness.

## Response Format
Return valid XML using exactly this structure:
<evaluation>
  <report>
    <informativeness_and_depth><score>X</score><justification>...</justification></informativeness_and_depth>
    <coherence_and_organization><score>X</score><justification>...</justification></coherence_and_organization>
    <verifiability><score>X</score><justification>...</justification></verifiability>
    <visualization_quality><score>X</score><justification>...</justification></visualization_quality>
    <original_image_integration><score>X</score><justification>...</justification></original_image_integration>
  </report>
</evaluation>

\textbf{User:}
## Topic
\{topic\}

<report>
\{report\}
</report>

The following images are original/source images inserted into the report.
\{source\_images\}

The following images are generated charts or rendered report visuals, not original/source images.
\{generated\_images\}
\end{Verbatim}
\end{tcolorbox}

\newpage

\end{document}